\documentclass[runningheads]{llncs}

\def\doi#1{\href{https://doi.org/\detokenize{#1}}{\url{https://doi.org/\detokenize{#1}}}}
\usepackage{graphicx}
\usepackage{listings}
\usepackage{amsmath}
\usepackage{amsfonts}
\usepackage{booktabs}
\usepackage{subcaption}
\usepackage{xspace}
\usepackage{siunitx}
\usepackage{pgfplots}
\usepackage{tikz}
\usepackage{pstricks}
\usepackage[misc,geometry]{ifsym} 

\usepackage[hyphens]{url}
\usepackage[pagebackref=true,breaklinks=true]{hyperref}
\hypersetup{
     colorlinks,
     linkcolor={red!75!black},
     citecolor={blue!75!black},
     urlcolor={blue!75!black},
}
\usepackage[capitalise]{cleveref}

\usepackage[inline]{enumitem}
\setlist*[enumerate]{label=(\arabic*)}

\makeatletter
\DeclareRobustCommand\onedot{\futurelet\@let@token\@onedot}
\def\@onedot{\ifx\@let@token.\else.\null\fi\xspace}

\def\eg{\emph{e.g}\onedot} 
\def\ie{\emph{i.e}\onedot}

\def\etal{\emph{et al}\onedot}
\makeatother

\usepackage{fancyhdr}
\fancypagestyle{specialfooter}{%
  \fancyhf{}
  
  \fancyfoot[L]{\scriptsize{$\copyright$ International Conference on Medical Image Computing and Computer Assisted Intervention (MICCAI) 2022, scheduled for September 18-22, 2022 in Singapore.}}
}
\begin{document}
\title{Multi-modal Retinal Image Registration Using a Keypoint-Based Vessel Structure Aligning Network}
\titlerunning{Multi-modal Retinal Image Registration}
%
 \author{Aline Sindel\inst{1} \and
 Bettina Hohberger\inst{2} \and
 Andreas Maier\inst{1} \and
 Vincent Christlein\inst{1}} 
\authorrunning{A. Sindel et al.}
%
\institute{Pattern Recognition Lab, FAU Erlangen-N\"urnberg, Erlangen, Germany \\
\email{aline.sindel@fau.de} \and
Department of Ophthalmology, Universit\"atsklinikum Erlangen, Erlangen, Germany}
\maketitle              
%
\thispagestyle{specialfooter}
\begin{abstract}
In ophthalmological imaging, multiple imaging systems, such as color fundus, infrared, fluorescein angiography, optical coherence tomography (OCT) or OCT angiography, are often involved to make a diagnosis of retinal disease. Multi-modal retinal registration techniques can assist ophthalmologists by providing a pixel-based comparison of aligned vessel structures in images from different modalities or acquisition times. To this end, we propose an end-to-end trainable deep learning method for multi-modal retinal image registration. Our method extracts convolutional features from the vessel structure for keypoint detection and description and uses a graph neural network for feature matching. The keypoint detection and description network and graph neural network are jointly trained in a self-supervised manner using synthetic multi-modal image pairs and are guided by synthetically sampled ground truth homographies. Our method demonstrates higher registration accuracy as competing methods for our synthetic retinal dataset and generalizes well for our real macula dataset and a public fundus dataset.

\keywords{Multi-modal retinal image registration  \and Convolutional neural networks \and Graph neural networks.}
\end{abstract}
\section{Introduction}
For the diagnosis of retinal disease, such as diabetic retinopathy, glaucoma, or age-related macular degeneration, and for the long-term monitoring of their progression, ophthalmological imaging is essential. Images are recorded over varying time periods using different multi-modal imaging systems, such as color fundus (CF), infrared (IR), fluorescein angiography (FA), or the more recent optical coherence tomography (OCT) and OCT angiography (OCTA). For the comparison and fusion of the information from different images by the ophthalmologists, multi-modal image registration is required to accurately align the vessel structures in the images. 

Multi-modal retinal registration methods can be summarized into global methods to predict an affine transform or a homography and local methods that estimate a non-rigid displacement field. In this work, we concentrate on feature-based methods that apply keypoint detection, description, matching, and computation of the global transform.
Classical methods estimate \eg the partial intensity invariant feature descriptor (PIIFD)~\cite{ChenJ2010} and Harris corner detector. This was extended by~\cite{WangG2015} using speed up robust feature (SURF) detector, \mbox{PIIFD}, and robust point matching, called SURF–PIIFD–RPM.
With the use of deep learning, some steps or even all steps are replaced by neural networks.
The retinal method DeepSPA~\cite{LeeJ2019} uses a convolutional neural network (CNN) to  
classify patches of vessel junctions based on a step pattern representation.
The keypoint detection and description network RetinaCraquelureNet~\cite{SindelA2022} is trained on small multi-modal retinal image patches centered at vessel bifurcations and uses mutual nearest neighbor matching and random  sample  consensus (RANSAC)~\cite{FischlerMA1981} for homography estimation.
In GLAMpoints~\cite{TruongP2019}, homography guided self-supervised learning is applied to train a UNet 
~\cite{RonnebergerO2015} for keypoint detection combined with RootSIFT~\cite{ArandjelovicR2012} descriptors for retinal image data.
The weakly supervised method by Wang \etal~\cite{WangY2021} sequentially trains a vessel segmentation network using style transfer and the mean phase image as guidance, the  self-supervised SuperPoint~\cite{DeToneD2018} network, and an outlier network using context normalization~\cite{YiKM2018}, which they adapt for the homography estimation task. 
End-to-end networks are often designed to directly compute the parameters of the transform. To predict affine and non-rigid transforms, there is for instance the image and spatial transformer networks (ISTN) for structure-guided image registration that learns a representation of the segmentation maps during training~\cite{LeeM2019}. 
An approach~\cite{ArarM2020} on spatial transformers and CycleGANs~\cite{ZhuJY2017} for multi-modal image registration uses cross-modality translation between the modalities to employ a mono-modality metric.

\begin{figure}[t]
	\centering
\includegraphics[width=\textwidth]{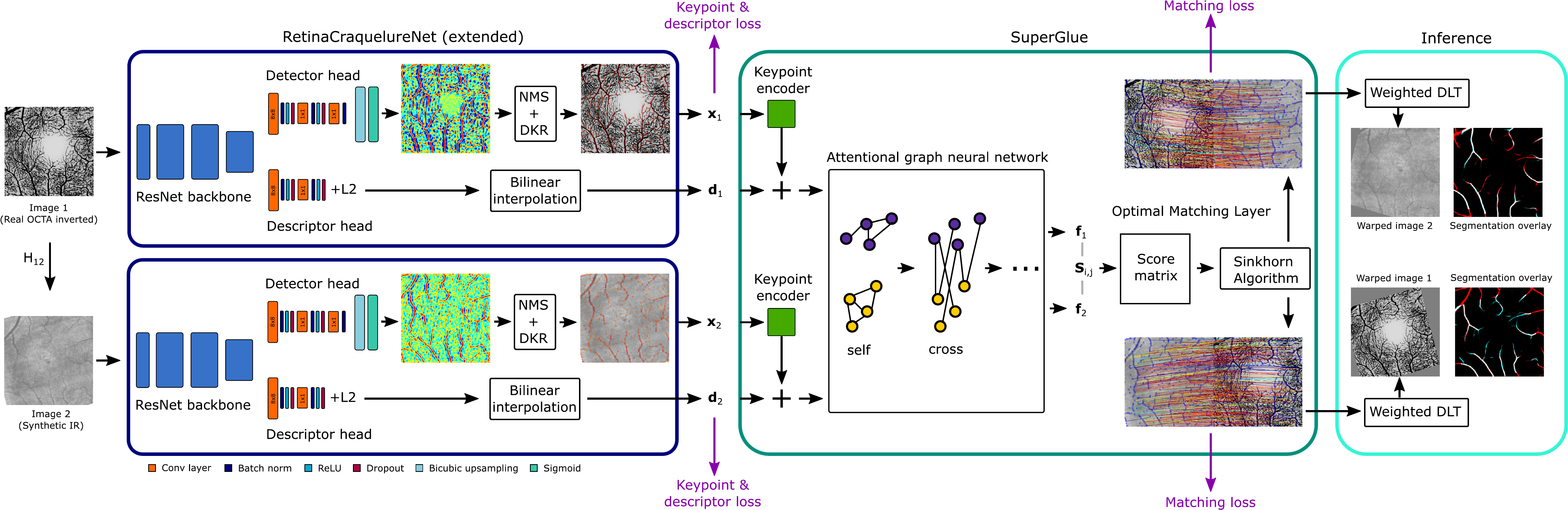}
	\caption{Our keypoint-based vessel structure aligning network (KPVSA-Net) for multi-modal retinal registration uses a CNN to extract cross-modal features of the vessel structures in both images and a graph neural network for descriptor matching. Our method is end-to-end and self-supervisedly trained by using synthetically augmented image pairs. During inference, the homography is predicted based on the matches and scores using weighted direct linear transform (DLT).}
	\label{fig-01}
\end{figure} 

In this paper, we propose an end-to-end deep learning method for multi-modal retinal image registration, named \textit{Keypoint-based Vessel Structure Aligning Network} (KPVSA-Net). We employ prior knowledge by extracting deep features of the vessel structure using the keypoint detection and description network RetinaCraquelureNet~\cite{SindelA2022}. In contrast to vessel segmentation based methods, we extract the features directly from multi-modal images to learn distinctive cross-modal descriptors. We build an end-to-end network for feature extraction and matching by extending RetinaCraquelureNet and combining it with the graph neural network SuperGlue~\cite{SarlinPE2020}. We jointly train both networks using a novel self-supervised keypoint and descriptor loss and a self-supervised matching loss guided by sampled homographies. We created a synthetically augmented dataset by training an image translation technique to generate synthetic retinal images. Our network incorporates and connects knowledge about the local and global position, visual appearance, and context between keypoints showing high registration accuracy. 
\begin{figure}[t]
	\centering		
	\includegraphics[width=.9\textwidth]{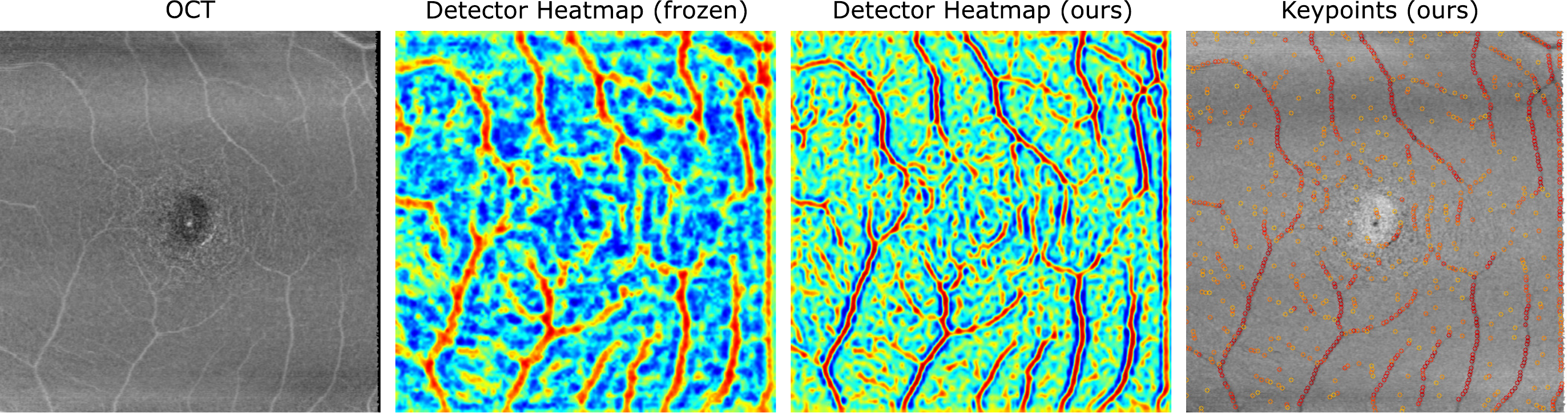}
	\caption{Keypoint confidence heatmap (from low confidence blue to high confidence red) without (middle left) and with (middle right) our novel self-supervised keypoint and descriptor loss in combination with the differentiable keypoint refinement. The extracted keypoints (most right) of our multi-modal registration method are color-coded based on their confidence (red is high).}	
	\label{fig-02}
\end{figure}

\section{Methods}
\subsection{Synthetic Augmentations for Multi-modal Retinal Images}
Our proposed method is trained end-to-end in a self-supervised manner guided only by synthetically sampled ground truth homographies. To apply the self-supervised technique to multi-modal images, we make use of unpaired image-to-image translation using the cycle consistency~\cite{ZhuJY2017}. 
For each modality combination, we train one CycleGAN~\cite{ZhuJY2017} to augment the training dataset by generating synthetic images of the other modalities.
To train our registration method, we sample random homographies on the fly to transform the second image. Afterwards, we crop both images at the same randomly selected position to a fixed patch size and recalculate the homographies based on the new corner points. We augment both patches independently with photometric augmentations such as color jittering, Gaussian blurring, sharpening, Gaussian random noise, and small random crops. Prior to warping, we jointly augment the full-size images with geometric transformations such as horizontal and vertical flipping, rotation, and elastic deformation by random noise. 

\subsection{Multi-modal Retinal Keypoint Detection and Description Network}
We employ and extend the fully-convolutional RetinaCraquelureNet~\cite{SindelA2021,SindelA2022} for our end-to-end pipeline (see~\cref{fig-01}).
The network architecture is composed of a ResNet~\cite{HeK2016} backbone and a keypoint detection and description head. The keypoint detection head has two output channels (``vessel'', ``background''), which we reduce to only one channel to directly predict the keypoint confidence score. We set the feature dimension of the description head to 256-D to reduce the parameters for end-to-end learning.
We pretrain the network from scratch using multi-modal retinal image patches centered at supervised keypoint positions with a binary cross-entropy loss for keypoint detection and a cross-modal bidirectional quadruplet descriptor loss~\cite{SindelA2021,SindelA2022}. 

\begin{figure}[t]
	\centering			
	\includegraphics[width=.95\textwidth]{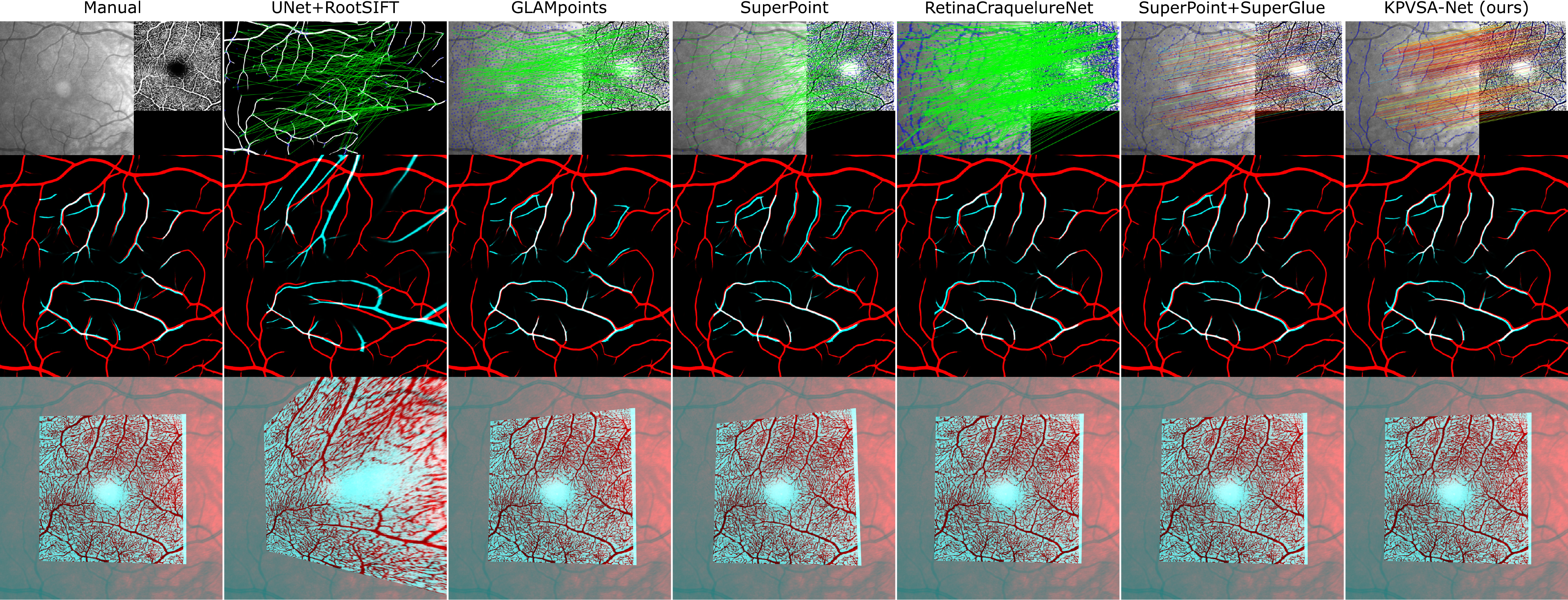}	
	\caption{Qualitative results for one IR-OCTA example.}	
	\label{fig-03}
\end{figure}

Then, we fit the network into our pipeline. In order to directly use the output of the detection head as keypoint confidence scores, we add a batch normalization layer after the last $1 \times 1$ convolutional layer and add a sigmoid activation after the bicubic upsampling layer. With these modifications the predictions of the detector are scaled to the range zero to one. 
Then, we apply non-maximum suppression (NMS) to the keypoint confidence heatmap and extract the top $N_\text{max}$ keypoints from the NMS heatmap~\cite{SindelA2021,SindelA2022}. This step is non-differentiable, therefore we apply a differentiable subpixel keypoint refinement (DKR) that allows the gradients to flow back to the small regions around the keypoints. Inspired by recent works~\cite{NibaliA2018,OnoY2018,JauYY2020,ZhaoX2022}, we extract $5 \times 5$ patches from the confidence heatmap which  are centered at the $N_\text{max}$ keypoint positions and compute for each patch $p$ the spatial softargmax of the normalized patch
 $(p - s_\text{NMS})/t$, where $s_\text{NMS}$ is the value of the NMS score map and $t$ the temperature for the softmax. The refined keypoint positions are the sum of the initial coordinates and the soft subpixel coordinates. 
The corresponding descriptors are linearly interpolated at the refined keypoint coordinates~\cite{SindelA2021}.

Based on the idea of the bidirectional quadruplet descriptor loss ($\mathcal{L}_{\text{Desc}}$)~\cite{SindelA2021}, we design a self-supervised keypoint and descriptor loss ($\mathcal{L}_{\text{KD}}$) that is guided by ground truth homographies instead of labeled matching keypoint pairs as in~\cite{SindelA2021}.
Within the detected keypoints in both images, positive keypoint pairs are automatically determined based on mutual nearest neighbor matching of the keypoint coordinates whose reprojection error is smaller than a threshold $\tau$. For the positive descriptor pairs (anchor $\mathbf{d}_a$ and positive counterpart $\mathbf{d}_p$), the closest non-matching descriptors in both directions are selected analogously to~\cite{SindelA2021}:
\begin{align}
\begin{split}
\mathcal{L}_{\text{Desc}}(\mathbf{d}_a,\mathbf{d}_p,\mathbf{d}_u,\mathbf{d}_v) = &\max [0, m + D(\mathbf{d}_a,\mathbf{d}_p) - D(\mathbf{d}_a,\mathbf{d}_u)]\\
+ &\max [0, m + D(\mathbf{d}_p,\mathbf{d}_a) - D(\mathbf{d}_p,\mathbf{d}_v)],
\end{split}
\end{align}
where $m$ is the margin, $D(x,y)$ the Euclidean distance, $\mathbf{d}_u$ the closest negative to $\mathbf{d}_a$, and $\mathbf{d}_v$ is the closest negative to $\mathbf{d}_p$.
However, since this self-supervised descriptor loss formulation depends on the number of matchable keypoints in the images with a coordinate distance smaller than $\tau$, it could encourage the reduction of the number of positive pairs $N_p$ to minimize the loss. 
To account for that and to refine the keypoint positions, we include a term into our loss to also minimize the reprojection error of the coordinates of the positive pairs (anchor $\mathbf{x}_a$, and warped coordinates of the positive counterpart $\hat{\mathbf{x}}_p$) which is weighted by $\beta$. This leads to our self-supervised keypoint and descriptor loss: 
\begin{align}
\begin{split}
\mathcal{L}_{\text{KD}}(\mathbf{d}_a,\mathbf{d}_p,\mathbf{d}_u,\mathbf{d}_v, \mathbf{x}_a, \hat{\mathbf{x}}_p) = \frac{1}{N_p}\sum_i^{N_p} \mathcal{L}_{\text{Desc}}(\mathbf{d}_{ai},\mathbf{d}_{pi},\mathbf{d}_{ui},\mathbf{d}_{vi})\\  
+  \frac{\beta}{N_p^2}\sum_i^{N_p} D(\mathbf{x}_{ai},\hat{\mathbf{x}}_{pi}).
\end{split}
\end{align}

\begin{table}[t]
\centering
\caption{Quantitative evaluation for our synthetic retina test dataset.  Models with * are fine-tuned on our synthetic augmented dataset.} 
\label{tab-01}
\tiny
\begin{tabular*}{\textwidth}{l@{\extracolsep\fill}rrrrrr@{\hspace{8pt}}rr}
\toprule
Metrics  & \multicolumn{6}{c}{Success rate for MHE ($\text{SR}_\text{MHE}$) [\%] $\uparrow$} & 
$\text{MHE}$ $\downarrow$ &  Dice $\uparrow$\\
 &$\epsilon=1$ & $\epsilon=2$ & $\epsilon=3$ & $\epsilon=4$ & $\epsilon=5$ & $\epsilon=10$ & Mean$\pm$Std & Mean$\pm$Std\\
\midrule
Before Reg & 0.0 & 0.0 & 0.0 & 0.0 & 0.0 & 0.0 & 78.10$\pm$36.44 & .083$\pm$.04 \\ 
UNet+RootSIFT & 7.6 & 20.1 & 27.1 & 32.4 & 35.8 & 42.2 & 368.93$\pm$2739.36 & .433$\pm$.34 \\ 
GLAMpoints* & 14.9 & 44.7 & 60.9 & 70.2 & 75.5 & 82.9 & 95.07$\pm$812.65 & .639$\pm$.25 \\ 
SuperPoint* & 11.0 & 33.3 & 55.6 & 67.7 & 76.6 & 90.7 & 8.61$\pm$53.55 & .709$\pm$.16 \\ 
RCN (512-D) & 13.7 & 43.1 & 57.8 & 65.4 & 69.0 & 73.2 & 107.70$\pm$1055.37 & .596$\pm$.31 \\ 
(SP+SG)* & 35.2 & 74.4 & 91.6 & 96.4 & 98.3 & 99.5 & 1.84$\pm$4.48 & .775$\pm$.11 \\ 
(SP*+SG)* & 39.9 & 79.5 & 92.9 & 97.4 & \textbf{99.1} & \textbf{99.7} & 1.49$\pm$1.50 & .781$\pm$.10 \\ 
\midrule
RCN-D*+SG*  & 47.4 & 88.2 & 95.1 & 97.0 & 98.2 & 99.0 & 1.46$\pm$2.51 & .783$\pm$.11 \\ 
RCN-KD*+SG* & 50.4 & 90.2 & 95.1 & 97.3 & 98.1 & 99.2 & 1.32$\pm$1.55 & .783$\pm$.11 \\ 
RCN-DK*-D*+SG* & 55.1 & 89.9 & 96.1 & 97.9 & 98.7 & 99.4 & \textbf{1.29}$\pm$2.13 & .782$\pm$.11 \\ 
KPVSA-Net & \textbf{74.2} & \textbf{94.9} & \textbf{98.1} & \textbf{98.6} & 98.7 & 99.1 & 1.36$\pm$6.45 & \textbf{.789}$\pm$.11 \\ 
\bottomrule
\end{tabular*}
\end{table}

\subsection{Keypoint Matching Using a Graph Convolutional Neural Network}
For keypoint matching, we incorporate the graph convolutional neural network SuperGlue~\cite{SarlinPE2020} into our method which consists of three building blocks, see~\cref{fig-01}. The keypoint coordinates are encoded as high dimensional feature vectors using a multilayer perceptron, and a joint representation is computed for the descriptors and the encoded keypoints~\cite{SarlinPE2020}. 

The attentional graph neural network (GNN) uses alternating self- and cross-attention layers to learn a more distinctive feature representation. 
The nodes of the graph are the keypoints' representations of both images. The self-attention layers connect the keypoints within the same image, while the cross-attention layers connect a keypoint to all keypoints in the other image.
Information is propagated along both the self- and cross-edges via messages. At each layer the keypoints' representations for each image are updated by aggregation of the messages using multi-head attention~\cite{VaswaniA2017}. 
Lastly, a $1 \times 1$ convolutional layer is used to obtain the final descriptors~\cite{SarlinPE2020}.

The optimal matching layer is used to compute the partial soft assignment matrix $\mathbf{P}$, which assigns for each keypoint at most one single keypoint in the other image.
Based on the score matrix of the similarity of the descriptors, $\mathbf{P}$ is iteratively solved using the differentiable Sinkhorn algorithm~\cite{SinkhornR1967}. To account for unmatchable keypoints, a dustbin is added to the $N \times M$ score matrix~\cite{SarlinPE2020}.
The negative log-likelihood of $\mathbf{P}$ is minimized~\cite{SarlinPE2020}:
\begin{equation}
\mathcal{L}_{\text{SG}}(\mathbf{P},\mathcal{M},\mathcal{I},\mathcal{J}) = - \kappa \sum_{(i,j) \in \mathcal{M}} \log \mathbf{P}_{i,j} - \sum_{i \in \mathcal{I}} \log \mathbf{P}_{i,N+1} - \sum_{j \in \mathcal{J}} \log \mathbf{P}_{M+1,j},
\end{equation}
where $\kappa$ is the weight for the positive matches $\mathcal{M}$, $\mathcal{I}$ denotes the unmatchable keypoints of image $I_I$, and $\mathcal{J}$ the unmatchable keypoints of image $I_J$ which are all those whose reprojection errors are higher than $\tau$.
We compute the $\mathcal{L}_{\text{SG}}$ and the ground truth matches twice, once based on matching from image $I_I$ to $I_J$ and once vice versa, \ie the matching loss is the sum of both. 

\begin{table}[t]
\centering
\caption{Quantitative evaluation for the IR-OCT-OCTA dataset. Models with * are fine-tuned on our synthetic augmented dataset.} 
\label{tab-02}
\tiny
\begin{tabular*}{\textwidth}{lrrrrrrrr@{\hspace{8pt}}rrrr}
\toprule
 & \multicolumn{2}{c}{IR-OCT} & \multicolumn{2}{c}{IR-OCTA} & \multicolumn{2}{c}{OCT-OCTA} & \multicolumn{2}{c}{All} &  & \multicolumn{3}{c}{All} \\
\cmidrule{2-9}
\cmidrule{11-13}
Metrics  & \multicolumn{8}{c}{Success Rates (ME<=7, MAE<=10) [\%] $\uparrow$} & & ME $\downarrow$ & MAE $\downarrow$ & Dice $\uparrow$ \\
& ME & MAE & ME & MAE & ME & MAE & ME & MAE && Mean$\pm$Std & Mean$\pm$Std & Mean$\pm$Std \\
\midrule
Before Reg & 0.0 & 0.0 & 0.0 & 0.0 & 30.0 & 26.7 & 10.0 & 8.9 && 123.89$\pm$79.25 & 128.59$\pm$79.65 & .117$\pm$.13 \\ 
Manual & 100.0 & 100.0 & 100.0 & 100.0 & 100.0 & 100.0 & 100.0 & 100.0 && 1.95$\pm$0.73 & 3.13$\pm$1.12 & .481$\pm$.09 \\ 
\midrule
UNet+RootSIFT & 13.3 & 6.7 & 13.3 & 8.3 & 73.3 & 65.0 & 33.3 & 26.7 && 210.09$\pm$452.84 & 640.46$\pm$2166.14 & .300$\pm$.22 \\ 
GLAMpoints* & 40.0 & 11.7 & 61.7 & 25.0 & \textbf{100.0} & 85.0 & 67.2 & 40.6 && 8.43$\pm$9.62 & 20.53$\pm$31.13 & .456$\pm$.12 \\ 
SuperPoint* & 35.0 & 16.7 & 26.7 & 21.7 & \textbf{100.0} & 93.3 & 53.9 & 43.9 && 89.98$\pm$427.20 & 365.69$\pm$2292.21 & .405$\pm$.18 \\ 
RCN (512-D) & 78.3 & 41.7 & 81.7 & 46.7 & \textbf{100.0} & 96.7 & 86.7 & 61.7 && 4.63$\pm$2.41 & 10.53$\pm$8.03 & .534$\pm$.09 \\ 
(SP+SG)* & 86.7 & 80.0 & 80.0 & 78.3 & \textbf{100.0} & \textbf{100.0} & 88.9 & 86.1 && 19.14$\pm$111.81 & 66.99$\pm$611.31 & .506$\pm$.15 \\ 
(SP*+SG)* & \textbf{100.0} & 90.0 & 93.3 & 88.3 & \textbf{100.0} & \textbf{100.0} & 97.8 & 92.8& & 3.85$\pm$2.46 & 7.08$\pm$5.64 & .533$\pm$.10 \\ 
KPVSA-Net & 96.7 & \textbf{91.7} & \textbf{98.3} & \textbf{93.3} & \textbf{100.0} & \textbf{100.0} & \textbf{98.3} & \textbf{95.0} && \textbf{3.67}$\pm$2.97 & \textbf{6.88}$\pm$8.45 & \textbf{.542}$\pm$.10 \\ 
\bottomrule 
\end{tabular*}
\end{table}

\section{Experiments}
\subsection{Multi-modal Retinal Datasets}
For our IR-OCT-OCTA retinal dataset, provided by the Department of Ophthalmology, FAU Erlangen-N\"urnberg, the maculas of $46$ controls were measured by Spectralis OCT II, Heidelberg Engineering up to three times a day resulting in $134$ images per modality and $402$ images in total. The multi-modal image triplets consist of IR images ($768 \times 768$) and en-face OCT and OCTA projections of the SVP layer (Par off) of the macula (both $512 \times 512$). We split the images for each modality into training: $89$, validation: $15$, and test set: $30$.
Secondly, we split the public color fundus (CF: $576 \times 720\times 3$) and  fluorescein angiography (FA: $576 \times 720$) dataset~\cite{ShirinH2021,HosseinDataset} that consists of $29$ image pairs of controls and $30$ pairs of patients with diabetic retinopathy into training: $35$, validation: $10$, and test set: $14$. 
For our synthetic dataset, we generate $1119$ multi-modal pairs of real and synthetic images for training, $205$ for validation, and $386$ for testing.
Due to our self-supervised training, we do not need any annotations, hence we only annotated $6$ control points per image for the test sets.
OCT, OCTA, and FA images are inverted for our experiments to depict all vessels in dark.

\subsection{Implementation and Experimental Details}
KPVSA-Net is implemented in PyTorch and we use the Kornia framework~\cite{RibaE2019} for data augmentation, homography estimation using weighted direct linear transform (DLT), 
and image warping. To initialize both networks, we pretrain our adapted version of RetinaCraquelureNet (RCN: 256-D) from scratch (backbone + detection head: learning rate of $\eta=1\cdot10^{-3}$, $100$ epochs; complete network: $\eta=1\cdot10^{-4}$, $25$ epochs; for both: with early stopping and linear decay of $\eta$ to $0$ starting at $10$) and use the SuperGlue weights of the Outdoor dataset. 
Then, we train KPVSA-Net end-to-end using Adam solver for $100$ epochs with $\eta=1\cdot10^{-4}$ for SuperGlue and $\eta=1\cdot10^{-6}$ for the detector and descriptor heads of RCN (frozen ResNet backbone) and then decay $\eta$ linearly to zero for the next $400$ epochs with early stopping and a batch size of $8$. The keypoint and descriptor loss and matching loss are equally weighted,  
$m=1$, $\beta=300$, $t=0.1$, $\kappa=0.45$, $\tau=3$, training patch size of $384$, $N_\text{max} = 512$ (synthetic dataset) or $N_\text{max} = 1024$ (real datasets), and matching score threshold of $0.2$ for DLT. 

For the comparison, we used the original configuration of RetinaCraquelureNet (RCN: 512-D)~\cite{SindelA2022} and we fine-tuned SuperPoint~\cite{DeToneD2018} (SP*) and GLAMpoints~\cite{TruongP2019} with our synthetic multi-modal dataset by extending the training code of~\cite{JauYY2020,TruongP2019}.
Then, we jointly fine-tuned SuperGlue and the descriptors of the pretrained SuperPoint model (SP+SG)* for $100$ epochs using our synthetic dataset and training strategy. Likewise, we jointly fine-tuned SuperGlue and the SP* model (SP*+SG)*. 
For the feature-based comparison methods, we use $N_\text{max}$ of $2000$ or $4000$ (synthetic/real), mutual nearest neighbor matching and RANSAC~\cite{FischlerMA1981} (reprojection error of $5$) for homography estimation using OpenCV.
For vessel segmentation and Dice score computation, we trained a UNet with synthetically augmented multi-modal images using CycleGANs based on the CF images and manual segmentations of the HRF dataset~\cite{BudaiA2013,HRFDataset}. 
The registration success rate for the real datasets is computed for the mean Euclidean error ($\text{ME}$) and maximum Euclidean error ($\text{MAE}$) of $6$ manual target control points and warped source control points using the predicted homography and an error threshold $\epsilon$. 
For the synthetic dataset, we compute the success rate of the mean homography error ($\text{MHE}$)~\cite{DeToneD2018} for different $\epsilon$ based on warping the corner points of the source image using the ground truth and the predicted homography.

\subsection{Results}
The quantitative results of the synthetic dataset are summarized in Table~\ref{tab-01}. Our KPVSA-Net shows the highest success rates for homography estimation for low error thresholds and in total the highest Dice score of the registered images. For error thresholds larger than $4$, the two SuperPoint+SuperGlue variants show comparable results. All SuperGlue-based methods achieve higher scores than the feature-based methods that use RANSAC. The bottom rows of Table~\ref{tab-01} show our ablation study. 
First, RCN (256-D) with training only the descriptor head, without keypoint refinement, and without our novel loss variant in combination with SuperGlue (RCN-D*+SG*) already shows an improvement of 7.5\,\% compared to (SP*+SG)* for $\epsilon<=1$. Enabling the keypoint detector and descriptor to learn (RCN-KD*+SG*) improves further by 3\,\%, and using the differential keypoint refinement (DKR) (RCN-DK*-D*+SG*) achieves 5\,\% more, and finally our full method KPVSA-Net additionally achieves 19\,\% plus for $\epsilon<=1$. The high accuracy for low error thresholds could be seen as the effect of the combination of our novel loss and DKR that pulls matching keypoints and descriptors closer together. The effect of both terms on the keypoint heatmap is visualized in Fig.~\ref{fig-02}. The left heatmap of the frozen detector highlights the vessel structures. Adding the single described steps only marginally change the visual appearance of the heatmap. Our final model has a refining effect on the heatmap (right) by thinning the high response area (red). Further, our loss also had a positive effect on SuperGlue by speeding up the convergence of both losses.
\begin{table}[t]
\centering
\caption{Quantitative evaluation for the public CF-FA dataset. Models with * are fine-tuned on our synthetic augmented dataset.}
\label{tab-03}
\tiny
\begin{tabular*}{\textwidth}{l@{\extracolsep\fill}rrrrrrr}
\toprule
Metrics  & \multicolumn{2}{c}{$\text{SR}_\text{ME}$} [\%] $\uparrow$ & \multicolumn{2}{c}{$\text{SR}_\text{MAE}$} [\%] $\uparrow$ &
$\text{ME}$ $\downarrow$ & $\text{MAE}$ $\downarrow$ & Dice $\uparrow$\\
 &$\epsilon=2$ & $\epsilon=3$ & $\epsilon=3$ & $\epsilon=5$ & Mean$\pm$Std & Mean$\pm$Std & Mean$\pm$Std\\
\midrule
Before Reg & 0.0 & 0.0 & 0.0 & 0.0 & 52.12$\pm$42.26 & 61.07$\pm$42.17 & .122$\pm$.03 \\ 
Manual & 100.0 & 100.0 & 92.9 & 100.0 & 0.73$\pm$0.34 & 1.24$\pm$0.71 & .606$\pm$.08 \\ 
\midrule
UNet+RootSIFT & 64.3 & 85.7 & 71.4 & 85.7 & 2.93$\pm$4.49 & 6.84$\pm$14.20 & .643$\pm$.14 \\ 
GLAMpoints* & 25.0 & 64.3 & 21.4 & 57.1 & 3.34$\pm$2.36 & 6.42$\pm$5.21 & .567$\pm$.11 \\ 
SuperPoint* & 71.4 & 85.7 & 46.4 & 78.6 & 1.85$\pm$0.72 & 3.41$\pm$1.38 & .636$\pm$.09 \\ 
RCN (512-D) & 71.4 & \textbf{100.0} & 64.3 & 89.3 & 1.70$\pm$0.55 & 3.04$\pm$1.37 & .658$\pm$.09 \\ 
(SP+SG)* & 85.7 & \textbf{100.0} & 71.4 & \textbf{100.0} & 1.56$\pm$0.51 & 2.58$\pm$0.85 & .648$\pm$.10 \\ 
(SP*+SG)* & \textbf{92.9} & \textbf{100.0} & 85.7 & \textbf{100.0} & 1.55$\pm$0.46 & 2.53$\pm$0.86 & \textbf{.661}$\pm$.09 \\ 
KPVSA-Net & \textbf{92.9} & \textbf{100.0} & \textbf{92.9} & \textbf{100.0} & \textbf{1.50}$\pm$0.36 & \textbf{2.47}$\pm$0.67 & .659$\pm$.09 \\ 
\bottomrule 
\end{tabular*}
\end{table}

The evaluation results for our real IR-OCT-OCTA dataset is shown in Table~\ref{tab-02} and for the public dataset in Table~\ref{fig-03}.
For the single multi-modal domain pairs, the twice fine-tuned (SP*+SG)* model has a slightly higher success rate for IR-OCT, but our method is slightly better for IR-OCTA and OCT-OCTA and for the total dataset. Generally, the errors are a bit higher for the real dataset and the best Dice score (ours) only has 54.2\,\% instead of 78.9\,\% for the synthetic dataset, but good results are still achieved. Since there is no ground truth for the real dataset, some inaccuracies come from the manual control points and also due to small deformations in the vessels or motion artifacts. The registration task for the CF-FA dataset is less complex, resulting in smaller ME and MAE for all methods and relatively close results for RCN, (SP+SG)*, (SP*+SG)*, and our method. We also tested the conventional method SURF+PIIFD+RPM~\cite{WangG2015} using their Matlab implementation. Results are in the supplementary material, as it achieved bad results for CF-FA and failed for the IR-OCT-OCTA dataset.

A qualitative IR-OCTA registration result is shown in Fig.~\ref{fig-03}. RootSIFT applied to the vessel segmentation predicted by the UNet finds the least number of correct matches and does not predict an acceptable homography. GLAMpoints detects more keypoints and matches than SuperPoint, but their registration results are comparable. The matches of RetinaCraquelureNet are concentrated on vessel structures resulting in a more precise registration. SuperPoint+SuperGlue filters out most false matches, but only shows a small number of matches in total. Our KPVSA-Net, however, detects a larger number of strong matches and results in a sightly more accurate overlay of the segmented vessels.

\section{Conclusion}
Our method incorporates prior knowledge of the vessel structure into an end-to-end trainable pipeline for retinal image registration. Using a graph neural network for image matching, spatial and visual information is connected to form a more distinctive descriptor. In the evaluation, our method demonstrates high registration accuracy for our synthetic retinal dataset and generalizes well for our real clinical dataset and the public fundus dataset. As there are some small deformations of the vessels, which cannot be handled with a perspective transform, we will look into non-rigid approaches as a further step of investigation.

%
%
%
\bibliographystyle{splncs04}
\bibliography{refs}

\end{document}